# Proposing a two-step Decision Support System (TPIS) based on Stacked ensemble classifier for early and low cost (step-1) and final (step-2) differential diagnosis of Mycobacterium Tuberculosis from non-tuberculosis Pneumonia

Toktam Khatibi[1,2]*, Ali Farahani[3], Hossein Sarmadian[4]

1: **(corresponding author)** Assistant Professor, School of Industrial and Systems Engineering, Tarbiat Modares University (TMU), Tehran, Iran, 14117-13114, email: toktam.khatibi@modares.ac.ir, Phone: +982182883913

2: Assistant Professor, Hospital Management Research Center (HMRC), Iran University of Medical Sciences (IUMS), Tehran, Iran.

[3] Computational Analysis and Modeling, Louisiana Tech University, Ruston, LA, United States

[4] Assistant Professor, Arak University of Medical Science, Arak, Iran.

**Abstract**

Background: Mycobacterium Tuberculosis (TB) is an infectious bacterial disease presenting similar symptoms to pneumonia; therefore, differentiating between TB and pneumonia is challenging. Therefore, the main aim of this study is proposing an automatic method for differential diagnosis of TB from Pneumonia.

Methods: In this study, a two-step decision support system named TPIS is proposed for differential diagnosis of TB from pneumonia based on stacked ensemble classifiers. The first step of our proposed model aims at early diagnosis based on low-cost features including demographic characteristics and patient symptoms (including 18 features). TPIS second step makes the final decision based on the meta features extracted in the first step, the laboratory tests and chest radiography reports. This retrospective study considers 199 patient medical records for patients suffering from TB or pneumonia, which has been registered in a hospital in Arak, Iran.

Results: Experimental results show that TPIS outperforms the compared machine learning methods for early differential diagnosis of pulmonary tuberculosis from pneumonia with AUC of 90.26±2.30 and accuracy of 91.37±2.08 with 95% CI and final decision making with AUC of 92.81±2.72 and accuracy of 93.89±2.81 with 95% CI.

Conclusions: The main advantage of early diagnosis is beginning the treatment procedure for confidently diagnosed patients as soon as possible and preventing latency in treatment. Therefore, early diagnosis reduces the maturation of late treatment of both diseases.

## 1  Background:

Mycobacterium Tuberculosis (TB) is an infectious bacterial disease which most commonly affects the lungs (1). A person infected with TB bacteria may have no symptoms. Patients with active TB need long-course treatment. TB is one of the top-10 causes of mortality worldwide (2) and the leading cause of death from infectious diseases (3). In 2016, about 10.4 million people contracted TB and 1.8 million died from it (2).

TB can present a wide range of symptoms, the most common being cough, blood in the sputum, fever, weight loss, weakness, night sweats, and chest pains (1). Other symptoms of TB include chills, fatigue, malaise, swollen lymph nodes, shortness of breath, phlegm, and loss of appetite (1).

Early detection of TB is vital for effective treatment, higher survival rate, and preventing further transmission of mycobacterium tuberculosis. Sputum smear tests and many other diagnostic tools have been used for early diagnoses. Blood tests and sputum tests are tedious and take a long time for analysis (4) though they are not always both necessary.

Delays from healthcare systems and from obtaining laboratory results occur frequently. (5). Previous researchers believe that correct diagnosis of TB takes 45 days (6). Another study has reported that there the median delay of diagnosis has been 9.9 weeks for tuberculosis detection (7).

Some TB symptoms can occur in other illnesses which affect the lungs. Pneumonia is a lung infection caused by bacteria, viruses or fungi. It displays similar symptoms to pulmonary TB (PTB) (8). Infectious disease experts find through experimentation that TB and pneumonia are incorrectly diagnosed interchangeably. Pinto et al. have confirmed that misdiagnosis is common (8). This is a cause of difficulty indistinguishing TB and non-tuberculosis Pneumonia and can lead to delay of treatment. Delay of treatment leads to reduce the patient survival rate, increased treatment cost, and lengthened treatment time interval.

Previous studies have shown that artificial intelligence using machine learning algorithms can be helpful for accurately diagnosing TB as soon as possible (6, 9, 10). Machine learning provides useful tools for locating hidden patterns in data sets, which can lead to discovery of valuable knowledge.

Researchers have made remarkable efforts to predict tuberculosis or pneumonia as well as differentially diagnose TB. Machine learning methods such as artificial neural networks (6, 11-13), decision trees (6, 14, 15), ANFIS (6), K-nearest neighbors (16), Support vector machines (16), logistic regression (17) ,and random forests (18) are used to differentiate.

To the best of our knowledge, distinguishing TB and Pneumonia based on machine learning methods has not been considered in previous studies. Therefore, the main aim of this study is to propose a method for differential diagnosis of TB and pneumonia.

Moreover, most of the previous studies have trained single classifiers on the dataset to predict tuberculosis, pneumonia and/or differentiating between TB from other diseases. Single machine learning models may be good estimators of a subset of patients; however, overfitting or under-fitting may occur for them, which may reduce their generalization ability.

Ensemble classifiers can improve the prediction performance by combining the results of different single classifiers (19). Previous studies have used ensemble classifiers with different base classifiers to predict ovarian cancer (20), kidney stone type (21), molecular-level features of cancer (22), and so on.

A new generation of ensemble classifiers with promising results are stacked ensemble (SE) methods (23). They consist of multiple layers of ensembles and can improve the performance of classification significantly (24). SEs are based on meta-learners to learn from the output of the base classifiers and they provide meta learners higher generalization ability (23). Therefore, in this study, we propose a stacked ensemble classifier for differentiating between TB and Pneumonia and compare its performance with other common and popular ensemble classifiers including random forests (25), Adaboost (26) and Gradient Boosting tree (27), as well as single classifiers including decision trees (28), logistic regression, support vector machines (29) and K-nearest neighbors (30).

Different features have been considered in the previous studies for diagnosing tuberculosis or pneumonia including the results of preliminary examination and laboratory tests (11, 12), patient symptoms and TB risk factors (17), demographic features (gender and age), and radiological features extracted from CT-Scan images (16, 31, 32). Smear test is a good predictor for TB detection but some TB infected patients may have negative smear test (33).

In this paper, a decision support system named TPIS based on stacked ensemble model is proposed to distinguish TB from pneumonia in two steps: early diagnosis and final decision making. For this purpose, demographic characteristics and patient symptoms are used for early diagnosis. The results of the laboratory tests and chest X-ray radiology reports are used for final decision making. The steps of TPIS are detailed bellow:

- The first step aims at early diagnosis based on low-cost features including demographic characteristics, and patients' symptoms. For this purpose, a stacked ensemble classifier is designed.
- The second step makes the final decision based on the results of the laboratory tests, chest X-ray reports, and meta features extracted in the first step. In this step, an ensemble layer of the classifiers is appended to the first-step's stacked ensemble.

The main advantage of our proposed method for differential diagnosis is preventing delay of treatment. The patient's treatment can be started as soon as possible based on the early diagnosis of the first step of TPIS. The features used for early diagnosis are low-cost and can be measured by physicians in their offices quickly. After that, for more confidence, the laboratory tests and chest X-rays are requested for the potentially infected patients and final decision making will be based on the results of the laboratory tests, chest X-ray reports, and their early diagnosis meta features. The treatment procedure may be modified after final decision making for suspicious patients.

The main contributions of this study are outlined as:

- Proposing a DSS for early differential diagnosis of TB and pneumonia based on low-cost features without wasting time to wait for the results of the laboratory tests and preparing radiology reports for chest X-ray

- Proposing an approach to identify potentially infected patients and requesting the laboratory tests and chest X-rays to improve the accuracy of differential diagnosis

- Proposing a stacked ensemble method for early diagnosis based on demographic characteristics and patients' symptoms

- Proposing a stacked ensemble method for final decision making based on the results of the laboratory tests, chest X-ray reports, and meta features extracted by the first-step's method

The rest of the paper is organized as follows: In section 2 the data characteristics and the main steps of the proposed method in this research is described. The results of the experiment, including the performance measures of the model, are reported and compared with the other classifiers in section 3. Concluding remarks and future work opportunities are presented in section 4.

## 2  Methods

The main steps of analyzing data for differentiating between TB and Pneumonia is shown in Figure 1:

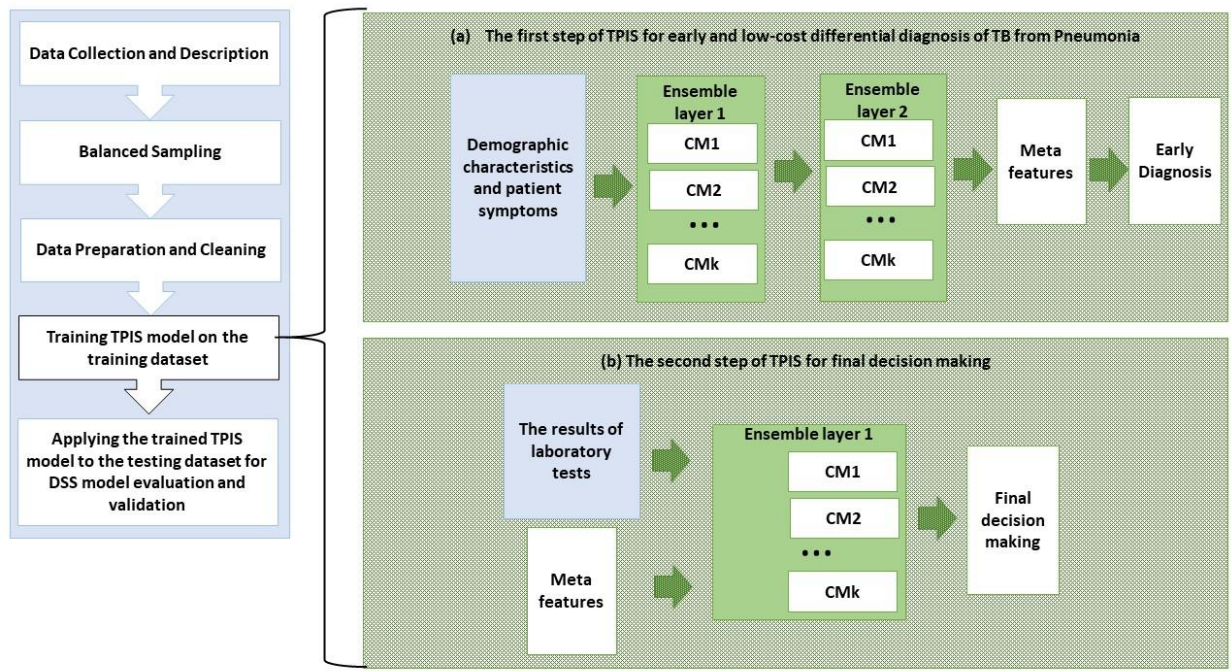

**Figure 1- (left): the main steps of this study methodology, (right): the proposed DSS (TPIS) framework**

The steps shown in Figure 1 (left) are based on the CRISP datamining methodology (34) which is one of the most powerful methods in datamining (21). The steps are described in more detail bellow.

- **Data collection and description**

For our collected data, disease type for each patient record is identified by a community of specialists based on sputum smear test, sputum culture and chest radiography. The patients with high disagreement degree among the specialists for their disease type are excluded from the study. After this filtering, 199 patients are considered in this study. Patient medical records for patients having TB and Pneumonia were collected from infectious unit of a hospital in Arak. Characteristics of the numeric features are listed in Table 1.

**Table 1- list of the numeric features extracted from the patients' medical records**

| Feature name | Total distribution | | | | | Distribution for patients suffering from pneumonia | | | | | Distribution for patients suffering from TB | | | | |
|---|---|---|---|---|---|---|---|---|---|---|---|---|---|---|---|
| | Min | Mean | Median | Max | Std | Min | Mean | median | Max | Std | Min | Mean | Median | Max | Std |
| Age | 15 | 61.93 | 67 | 102 | 23.33 | 15 | 61.11 | 64 | 102 | 23.77 | 17 | 63.04 | 68 | 94 | 0.23 |

| Feature name | Total distribution | | | | | Distribution for patients suffering from pneumonia | | | | | Distribution for patients suffering from TB | | | | |
|---|---|---|---|---|---|---|---|---|---|---|---|---|---|---|---|
| | Min | Mean | Median | Max | Std | Min | Mean | median | Max | Std | Min | Mean | Median | Max | Std |
| white blood cell (WBC) | 3.02 | 8.57 | 7.7 | 27.3 | 3.70 | 3.6 | 8.36 | 7.7 | 19.2 | 3.08 | 3.02 | 8.87 | 7.65 | 27.3 | 4.47 |
| Hemoglobin | 5/8 | 13/22 | 12/9 | 65/25 | 4/64 | 5/8 | 13/31 | 13/25 | 19/26 | 2/22 | 7/95 | 13/08 | 12/05 | 65/25 | 6/82 |
| Hematocrit | 22/2 | 40/71 | 40/1 | 151/66 | 10/52 | 22/2 | 40/75 | 40/96 | 55/7 | 5/69 | 28/2 | 40/65 | 38/82 | 151/66 | 15/14 |
| Neutrophil | 55/78 | 85/83 | 85/66 | 102/58 | 6/61 | 55/78 | 85/39 | 85/53 | 100/2 | 6/56 | 61/42 | 86/50 | 86/56 | 102/58 | 6/67 |
| Lymphocyte | 8 | 70/83 | 72 | 105/27 | 12/83 | 40 | 69/91 | 70 | 93 | 11/9 | 8 | 72/22 | 73/5 | 105/27 | 14/068 |
| Mean Corpuscular Volume (MCV) | -12/31 | 22/34 | 20 | 78/4 | 11/53 | 4 | 23/34 | 21/83 | 53 | 10/71 | -12/31 | 20/84 | 20 | 78/4 | 12/59 |
| C-reactive protein: (CRP) | -0/77 | 1/96 | 2 | 6/55 | 1/45 | -0/77 | 1/88 | 2 | 4/91 | 1/45 | -0/11 | 2/07 | 2 | 6/55 | 1/46 |
| Erythrocyte Sedimentation Rate (ESR) | -11/74 | 35/13 | 27 | 124 | 29/09 | -8/26 | 27/62 | 20 | 98 | 22/72 | -11/74 | 46/31 | 37 | 124 | 33/74 |

Table 2 lists the distribution of binary features considered in this study.

**Table 2- Binary features considered in this study**

| Feature name | Total | For patients suffering from pneumonia | For patients suffering from TB |
|---|---|---|---|
| Disease Type | 199 | 119 | 80 |
| Gender | Female: 91, Male: 108 | Female: 58, Male: 61 | Female: 33, Male: 47 |
| Cough | No: 7, Yes: 192 | No: 7, Yes: 112 | No: 0, Yes: 80 |
| Sputum | Yes: 153, No:46 | Yes: 87, No: 32 | Yes: 66, No:14 |
| Bloody Sputum | No: 185, Yes: 14 | No: 112, Yes: 7 | No: 73, Yes: 7 |
| Fever | No: 81, Yes: 118 | No: 47, Yes: 72 | No: 34, Yes: 46 |

| Feature name | Total | For patients suffering from pneumonia | For patients suffering from TB |
|---|---|---|---|
| Shaking | No: 126, Yes: 73 | No: 57, Yes:62 | No: 69, Yes: 11 |
| Smoking | No: 175, Yes: 24 | No: 103, Yes: 16 | No: 72, Yes: 8 |
| Joint Pain | No: 195, Yes: 4 | No: 119, Yes: 0 | No: 76, Yes: 4 |
| Edema | No: 177, Yes: 22 | No: 105, Yes: 14 | No: 72, Yes:8 |
| Asthma | No: 85, Yes: 114 | No: 45, Yes: 74 | No: 40, Yes: 40 |
| Diabetes | No: 167, Yes: 32 | No: 103, Yes: 16 | No: 64, Yes: 16 |
| Cyanosis | No: 195, Yes: 4 | No: 117, Yes: 2 | No: 78, Yes: 2 |
| Weight Loss | No: 171, Yes: 28 | No: 114, Yes: 5 | No: 57, Yes: 23 |
| Weakness | No: 155, Yes: 24 | No: 95, Yes: 24 | No: 60, Yes: 20 |
| Lung sound | Normal: 163, Abnormal: 36 | Normal: 118, Abnormal: 1 | Normal: 45, Abnormal: 35 |
| Dyspnea | No: 169, Yes: 30 | No: 100, Yes: 19 | No: 69, Yes: 11 |
| Orthopnea | No: 181, Yes: 18 | No: 103, Yes: 16 | No: 78, Yes: 2 |
| Lung abnormalities in CXR | No: 5, Yes: 174 | No:1, Yes: 109 | No: 4, Yes: 65 |
| White spots (infiltrates) in CXR | No: 9, Yes: 166 | No:1, Yes: 103 | No: 8, Yes: 63 |

For more convenience, a histogram of some binary features is shown in Figure 2.

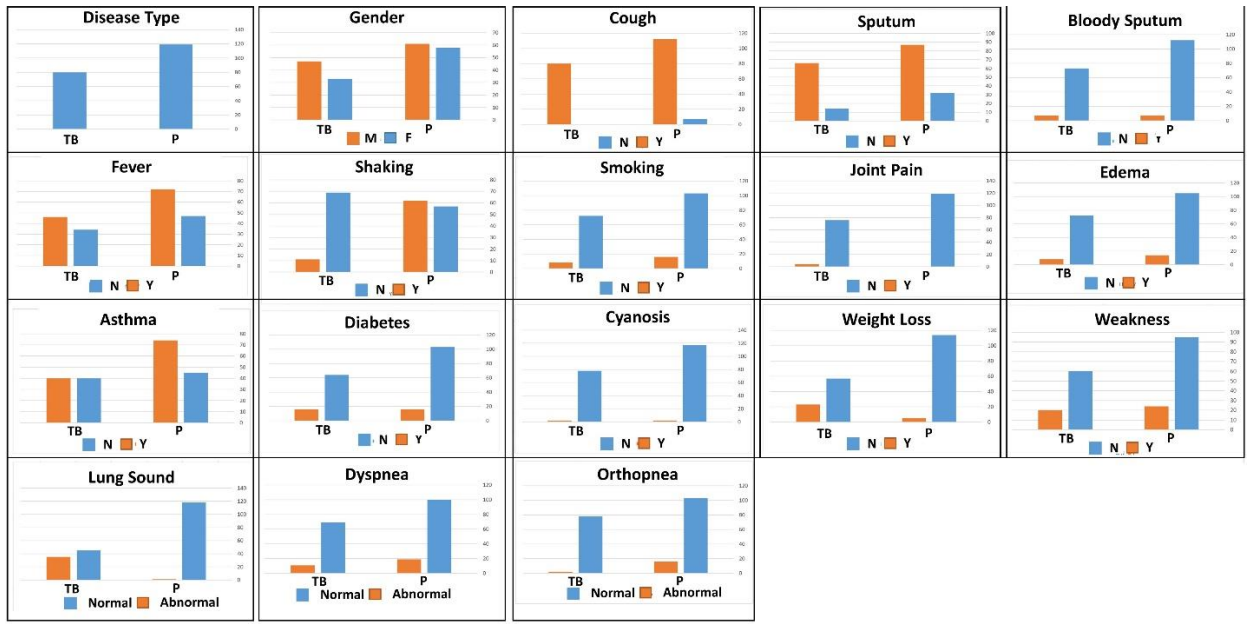

**Figure 2- histograms of some binary features**

As shown in Figure 22, when the features are considered individually, differentiation is difficult, and in each level of the binary feature, there are some patients suffering from either pneumonia or TB.

Two diagnostic tests for active tuberculosis or pulmonary tuberculosis are the sputum smear test and the sputum smear culture. But some patients having difficulties coughing up sputum. Moreover, sputum smear culture is a slow and expensive test, and its results depend on the quality of sputum (35). Many false negative results in the sputum smear test show its low efficiency for diagnosis of TB. Moreover, some infections such as HIV can lead to highly frequent negative-smear TB (36). Final diagnosis and disease type are performed by a community of specialists based on sputum smear test, sputum culture, chest radiography and clinical symptoms. The results of sputum smear test and smear culture are mainly used for differential diagnosis by specialists along with other descriptors and are not included in the input features.

- **Balanced Sampling**

For keeping and evaluating the generalization ability of the machine learning methods, the dataset should be split into non-overlapping training and testing datasets. Training and testing datasets will be used for training the classifiers and evaluating the trained classifiers, respectively. In this paper, balanced sampling (37) is performed for building the training dataset to avoid the majority class bias. Balanced sampling selects samples with equal size from each of the classes.

- **Data preparation and cleaning**

After collecting the data, it must be preprocessed in order to generate an appropriately prepared data set. The preprocessing tasks performed in this paper include outlier detection and removal, data transformation, and missing value imputation.

Outliers are detected by interaction with domain experts and boxplot analysis. Outliers are data points lying outside the whiskers of the boxplot (38).

Nominal features are converted to new binary variables, and ordinal and numerical features are normalized to [0,1] using the min-max normalization method. Without normalization, large-scaled features will dominate the others (38).

Features having high missing value rate (more than 30 percent) are excluded from the study. Missing value imputation for asymmetric binary features are performed by interaction with a domain expert. For other attributes, the K-nearest neighbor method is used for missing value imputation (39).

- **TPIS model training based on training dataset**

The main steps of TPIS are shown in Figure 1 (right). The first step aims at early diagnosis of TB from Pneumonia based on low cost features and assigns each patient a confidence score. After that, some laboratory tests and chest X-ray images are requested for patients with low confidence scores. Final diagnosis is performed based on their results in the second step of TPIS. More details on each step are described below.

## 2.1 The first step of TPIS for early differential diagnosis of TB from pneumonia based on low-cost features

The features contributing in the first step of TPIS as input variables include demographic characteristics (gender and age) and patient symptoms (having cough, fever, sputum, bloody sputum, shaking, joint pain, edema, asthma, diabetes, cyanosis, weight loss, weakness, dyspnea, orthopnea, normal or abnormal lung sound, and smoking or non-smoking). The considered symptoms are seen in patients suffering from TB or pneumonia frequently.

The input features of the first step of TPIS are fast and low cost because they can be described by the patient or determined and measured via a clinical examination in a physician's office. Therefore, their measurement has low financial cost and can be performed very quickly without spending time to do a laboratory test and obtaining the result.

The framework of the first step of TPIS with more details is shown in Figure 3.

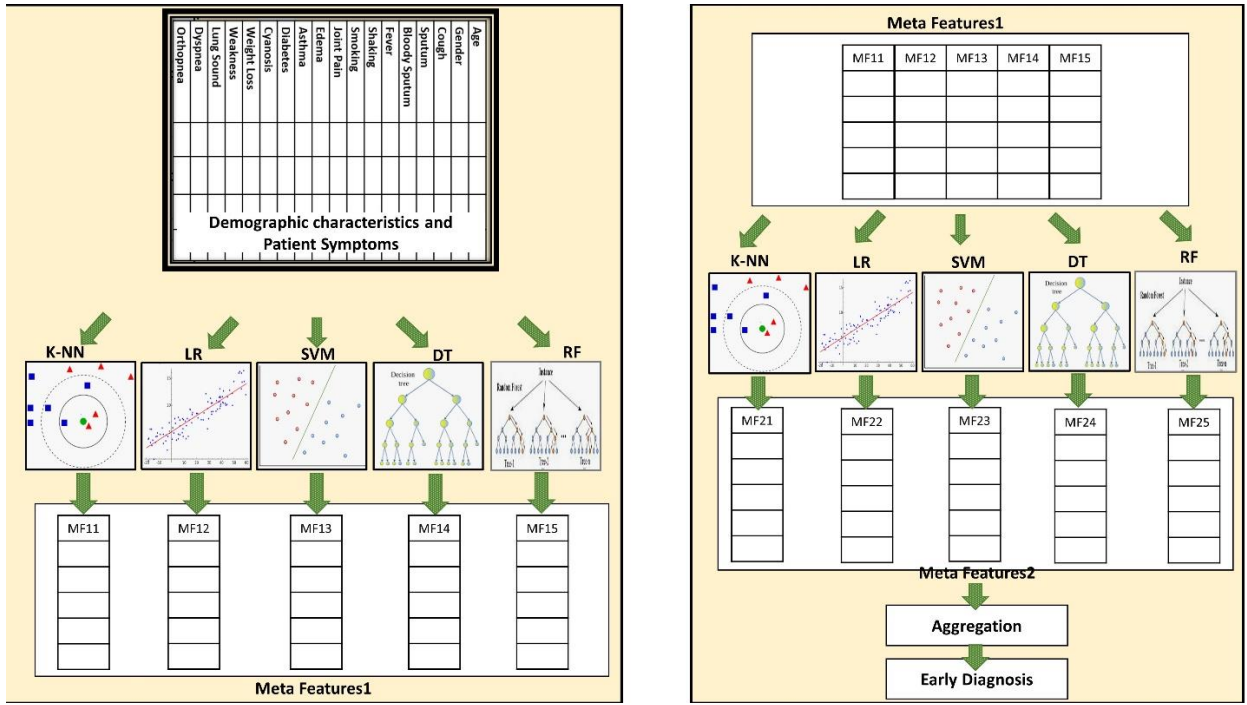

**Figure 3- the framework of the first step of TPIS with more details**

As shown in Figure 3, a stacked ensemble classifier is trained on the input features, which has two layers of ensemble methods. Previous studies have shown that the ensemble classifiers have higher accuracies when stacked together (23).

For choosing a good combination of the classifiers in the body of the ensemble classifiers, similarly to Kazemi and S. Mirroshandel (21), we have trained and evaluated several different classifiers. For this purpose, about 30% of the dataset is randomly selected as the development dataset, and the classifiers with the best performance on this dataset are selected. Both layers of our proposed SE in the first step use K-nearest neighbors (K-NN), Logistic regression (LR), support vector machines (SVM), decision trees (DT), and random forests (RF) as the base classifiers. They are implemented using the Python library Scikit-learn (version 0.21.1) (40).

The output of the base classifiers (K-NN, LR, SVM, DT, and RF) in each layer of SE are considered as the meta features. Meta features generated in the first layer (Meta Features 1) are fed into the second layer of SE as its input features.

In the second layer, after training the base classifiers on Meta Features 1, the outputs of them are aggregated based on a voting method and are used for early diagnosis.

In order to have a level of confidence in the decision made for each patient in the early diagnosis step, a confidence score is needed. For example, when all of the classifiers assign the same class label to a patient, disease diagnosis is performed with highest confidence. If only half of the base classifiers assign the same label, it shows the lowest confidence of disease detection for this patient.

Different confidence scores have been defined and used in the previous studies for semi-supervised classification. Unlabeled data with higher confidence scores are assigned a label and added to the training dataset (41). In this study, a strict confidence score (CS) is calculated for each patient's disease label as Eq. (1) The CS has a negative correlation with the level of disagreement among the base classifiers:

$$CS(P_i) = \left| \frac{\sum_{\{C_j : P_{C_j}(y=TB|P_i) > \varepsilon\}} 1 - \sum_{\{C_j : P_{C_j}(y=P|P_i) > \varepsilon\}} 1}{\sum_{\{C_j : P_{C_j}(y=TB|P_i) > \varepsilon\}} 1 + \sum_{\{C_j : P_{C_j}(y=P|P_i) > \varepsilon\}} 1} \right| \qquad \text{Eq. (1)}$$

Where $P_i$ is the patient i, $C_j$ is the $j^{th}$ base classifier, $P_{Cj}(y=TB|P_i)$ is the probability of assigning $P_i$ to class TB by $C_j$ and $P_{Cj}(y=P|P_i)$ is the conditional probability of classifying $P_i$ to pneumonia by $C_j$.

CS varies between 0 and 1. If all base classifiers assign the same class label to a patient, then CS will have the maximum value. The lowest CS is obtained when half of the base classifiers assign the same label to a patient.

We will apply a threshold to CS for identifying the patients with low confidence scores. These patients are named "potentially infected" patients in this paper. Some supplementary information is required to improve the confidence of disease diagnosis for patients. Therefore, some laboratory tests and chest X-rays are requested for them in the second step of TPIS.

## 2.2 The second step of TPIS for final decision making

The input features of the second step of TPIS include the results of the laboratory tests and the keywords extracted from chest X-ray reports. Laboratory tests include White blood cells (WBC), Hemoglobin, Hematocrit, Neutrophil, Lymphocyte, mean corpuscular volume (MCV), C-reactive protein (CRP), and Erythrocyte Sedimentation Rate (ESR).

The framework of the second step of TPIS with more details is shown in Figure 4.

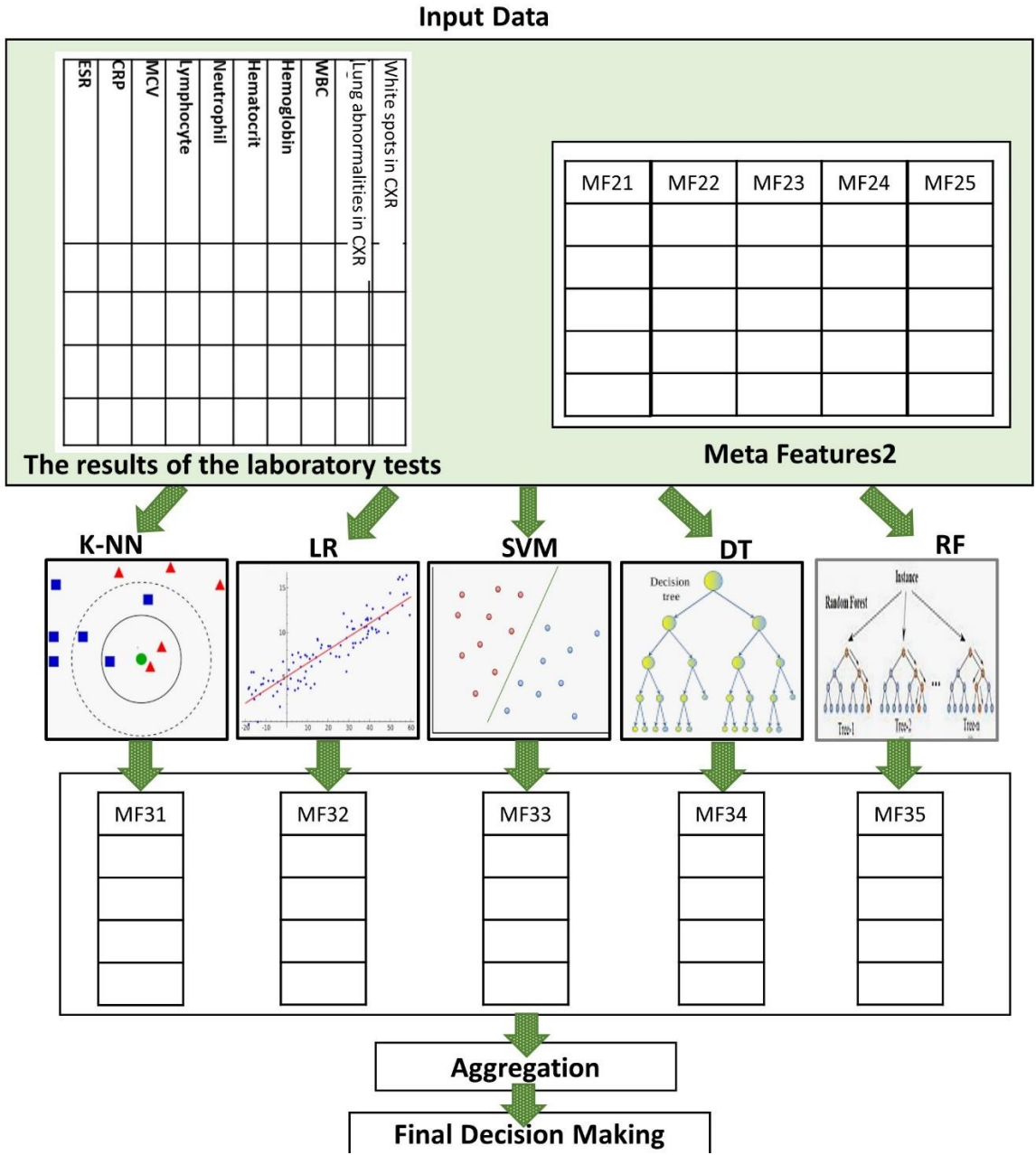

**Figure 4- the framework of the second step of TPIS with more details**

As shown in Figure 4, an ensemble classifier consisting of K-NN, LR, SVM, DT, and RF is trained in the second step. The input features include the results of the laboratory tests, the keywords extracted from chest X-ray reports, and meta features extracted from the second layer of the stacked ensemble model (Meta Features 2) in the first step of TPIS. Final decision making is performed by majority voting of the outputs of the base classifiers in this step.

## 3 Results

For evaluating and validating the proposed model for differential diagnosis of TB from pneumonia, some performance measurements should be calculated including Accuracy, Precision, Recall, F-Score, and Area under the ROC curve (AUC).

Accuracy calculates how many data records are classified correctly as Eq. (2):

$$\text{Accuracy} = \frac{TP + TN}{N} \qquad \text{Eq. (2)}$$

TP and TN are true positive and true negative data records and N is the number of all data records. Positive and negative classes are TB and pneumonia, respectively.

Precision measures the fraction of the real positive instances among the predicted positive ones as Eq.(3).

$$\text{Precision} = \frac{TP}{TP + FP} \qquad \text{Eq. (3)}$$

Recall denotes how many data records belonging to positive class are classified correctly as Eq. (4):

$$\text{Recall} = \frac{TP}{TP + FN} \qquad \text{Eq. (4)}$$

FP and FN are false positive and false negative data records.

F-Score is the harmonic average of Precision and Recall as Eq. (5):

$$F - \text{Score} = 2 \frac{\text{Precision} * \text{Recall}}{\text{Precision} + \text{Recall}} \qquad \text{Eq. (5)}$$

ROC (Receiver Operating Curve) is a probability graphical plot showing the discriminative power of the binary classifier by changing the threshold. AUC measures the area under ROC. It tells how much the model can distinguish between classes. The higher the AUC, the better the model is at predicting two classes.

For evaluating and validating our proposed method, different combinations of the demographic characteristics, patient symptoms, the results of the laboratory tests and X-ray reports, and the meta features extracted in the first step of TPIS are considered as shown in Table 3. Different feature sets have been used in this study for evaluation of the classifiers.

Table 3- description of feature sets

| Feature set name | Included features |
|---|---|
| $FS_1$ | Symptoms and demographic features |

| Feature set name | Included features |
|---|---|
| $FS_2$ | The results of the laboratory tests and the keywords extracted from radiology report on chest X-ray image |
| $FS_3$ | Meta features (MetaFeatures2) obtained by our proposed early diagnosis model |
| $FS_4$ | The union of $FS_2$ and $FS_3$ |
| $FS_5$ | The union of $FS_1$ and $FS_2$ |

The evaluation framework used in this study is shown in

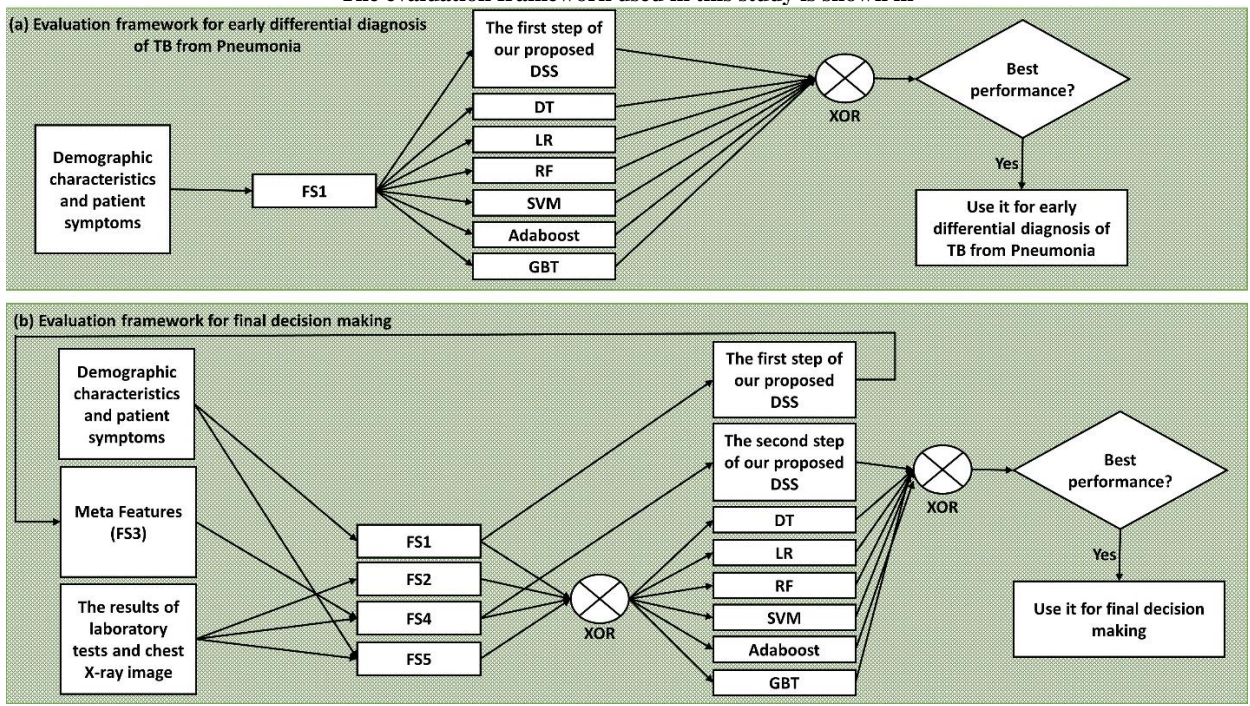

Figure 5.

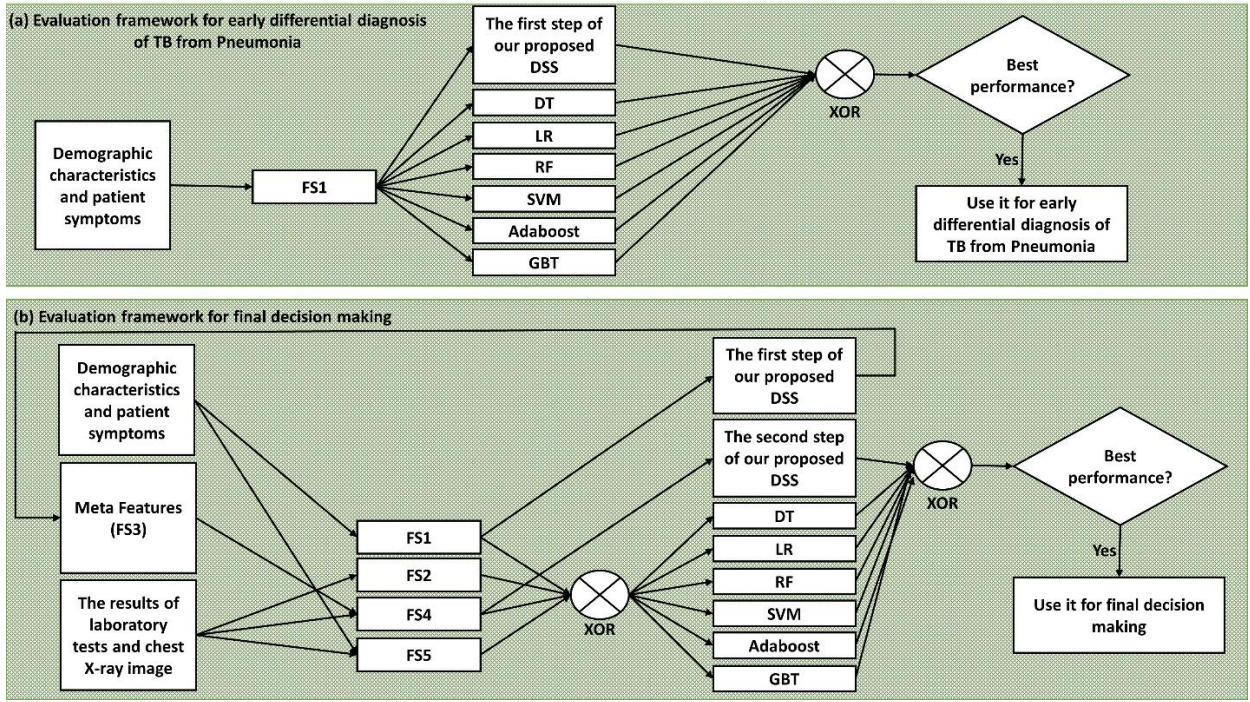

**Figure 5- the evaluation framework used in this study**

As shown in Figure 5, the evaluation framework used in this study consists of two modules, (a) and (b):

- (a) Evaluation framework for early differential diagnosis of TB and pneumonia:
    - Only demographic characteristics and patient symptoms are fed to the classifiers as input variables. The performance of the first step module of TPIS is compared with other single classifiers such as DT, LR and SVM, and ensemble methods such as RF. Adaboost were implemented using Python library Scikit-learn (version 0.21.1), and Gradient Boosting Tree (GBT) was implemented using Microsoft's software 'LightGBM' with a Python wrapper (42). The results of the comparison are listed in Table 4.
- (b) Evaluation framework for final decision making:
    - The second step of TPIS is trained by feeding it $FS_4$. $FS_4$ features consist of meta features ($FS_3$) extracted by the first step of TPIS, the results of the laboratory tests, and the keywords extracted from the radiology reports on chest X-ray images ($FS_2$). The performance of the second step of TPIS is compared to single classifiers such as DT, LR, and SVM, and ensemble methods such as RF, Adaboost, and GBT. The input features fed to the compared classifiers are the features of $FS_1$, $FS_2$, $FS_4$, or $FS_5$. The performance measures for these classifiers are listed in Table 5 and Table 6.

**Table 4- Comparing the performance of different classifiers for early differential diagnosis of TB and pneumonia with CI 95% based on symptoms and demographic features ($FS_1$)**

| Model | Accuracy | AUC | Precision | Recall | F-Score |
|---|---|---|---|---|---|
| Decision Tree | 89.00±2.69 | 84.58±2.58 | 80.23±2.41 | 94.17±2.36 | 86.64±2.42 |

| Model | Accuracy | AUC | Precision | Recall | F-Score |
|---|---|---|---|---|---|
| LR | 87.50±2.93 | 86.46±2.36 | 78.79±2.29 | 91.67±2.73 | 84.75±2.03 |
| SVM | 87.75±2.86 | 86.41±2.36 | 78.53±2.17 | 93.12±2.45 | 85.20±2.91 |
| RF | 89.5±2.85 | 88.85±3.96 | 82.30±2.40 | 92.08±2.32 | 86.92±2.43 |
| Adaboost | 86.12±2.05 | 85.74±2.88 | 76.25±2.52 | 92.01±2.27 | 83.39±2.35 |
| GBT | 85.87±2.43 | 84.65±2.97 | 76.20±2.66 | 91.18±2.47 | 83.02±2.52 |
| The first layer of the first step of TPIS | 90.62±2.19 | 89.58±2.32 | 83.61±2.79 | 95.23±2.70 | 88.49±1.79 |
| The second layer of the first step of TPIS | 91.37±2.08 | 90.26±2.30 | 83.61±2.61 | 96.04±2.27 | 89.39±2.56 |

As shown in Table 4, the first step of TPIS outperforms the compared classifiers for early disease diagnosis.

**Table 5- Comparing the performance of different classifiers for final differential diagnosis of TB and pneumonia with CI 95% based on the features in FS$_4$**

| Model | Accuracy | AUC | Precision | Recall | F-Score |
|---|---|---|---|---|---|
| Decision Tree | 92.54±1.15 | 91.45±1.37 | 88.10±1.46 | 92.84±2.17 | 90.41±2.02 |
| LR | 92.48±1.45 | 91.43±1.65 | 88.10±2.53 | 92.66±1.10 | 90.32±1.93 |
| SVM | 92.66±1.77 | 91.91±2.31 | 88.29±2.76 | 92.95±2.35 | 90.56±2.61 |
| Adaboost | 91.39±1.40 | 90.22±1.46 | 86.52±1.07 | 91.52±1.59 | 88.95±1.13 |
| GBT | 90.81±1.08 | 90.06±1.84 | 86.15±1.95 | 90.25±1.88 | 88.15±1.92 |
| RF | 92.67±2.12 | 91.87±2.40 | 88.42±2.08 | 92.80±1.97 | 90.56±2.21 |
| The second step of TPIS | 93.96±2.81 | 92.81±2.72 | 90.63±2.61 | 93.75±2.03 | 92.16±2.38 |

As illustrated by Table 5, the best performance for final decision making is obtained by TPIS. The classifiers trained on different feature sets are compared, as listed in Table 6.

**Table 6- Comparing the performance of different classifiers for final differential diagnosis of TB and pneumonia with CI 95% based on different features**

| Features | Model | Accuracy | AUC | Precision | Recall | F-Score |
|---|---|---|---|---|---|---|
| $FS_2$ | DT | 87.50±1.13 | 82.50±0.99 | 78.39±1.32 | 92.50±1.08 | 84.86±1.25 |
| | LR | 88.05±1.03 | 84.87±1.01 | 78.87±1.12 | 93.50±0.95 | 85.56±0.98 |
| | SVM | 88.20±1.26 | 84.87±1.22 | 79.57±1.26 | 92.62±1.33 | 85.60±1.27 |
| | Adaboost | 85.00±1.57 | 81.76±1.39 | 75.63±1.19 | 89.11±1.61 | 81.82±1.46 |
| | GBT | 84.43±0.92 | 80.88±0.88 | 75.98±1.17 | 88.32±0.72 | 81.69±1.04 |
| | RF | 89.00±1.18 | 85.04±1.18 | 81.48±1.22 | 91.83±1.06 | 86.35±1.12 |
| $FS_3$ | DT | 90.18±1.91 | 89.33±1.27 | 84.07±2.01 | 91.39±1.30 | 87.58±1.77 |
| | LR | 90.66±1.72 | 89.40±1.17 | 84.55±2.14 | 92.18±1.65 | 88.20±1.82 |
| | SVM | 91.45±1.59 | 89.69±2.42 | 85.66±1.85 | 93.00±1.54 | 89.18±1.59 |
| | Adaboost | 91.08±1.26 | 89.45±1.49 | 85.32±1.64 | 92.34±1.60 | 88.69±1.63 |
| | GBT | 90.02±1.55 | 89.21±1.24 | 83.27±2.46 | 92.17±1.41 | 87.49±1.78 |
| | RF | 91.03±1.08 | 89.63±1.51 | 84.61±2.01 | 93.28±1.18 | 88.74±1.59 |
| $FS_5$ | DT | 88.5±1.28 | 89.12±0.99 | 79.07±1.75 | 94.71±1.01 | 86.19±1.42 |
| | LR | 89.67±1.53 | 87.54±1.64 | 81.10±1.92 | 94.83±1.84 | 87.43±1.87 |
| | SVM | 90.00±1.03 | 90.09±1.62 | 80.60±1.33 | 91.83±0.90 | 85.85±1.08 |
| | Adaboost | 87.33±2.65 | 86.17±1.54 | 79.28±1.07 | 90.10±2.19 | 84.34±1.64 |
| | GBT | 85.5±2.14 | 84.45±2.18 | 76.24±3.22 | 89.67±1.25 | 82.41±2.52 |
| | RF | 90.02±1.23 | 89.75±2.07 | 83.31±2.30 | 92.11±0.80 | 87.49±1.58 |

Accuracy, AUC and F-Score of the compared methods are shown in Figure 6 for convenience.

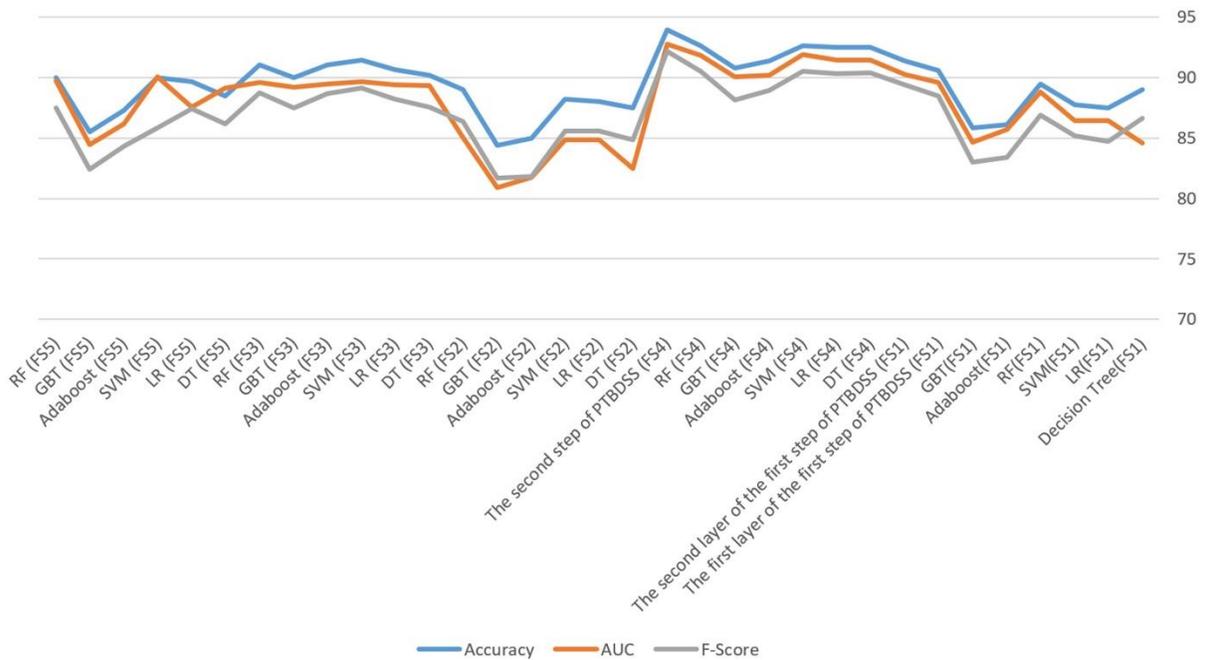

**Figure 6- Accuracy, AUC and F-Score of the compared classifiers**

As illustrated by Figure 6, the performance of the classifiers trained on $FS_2$ is good but measuring the laboratory tests requires time. On the other hand, radiologic report quality depends on the radiologists' expertise. Training the classifiers on $FS_5$ (including $FS_1$ and $FS_2$) improves their performance. It shows that $FS_5$ can predict TB and pneumonia highly accurately. Finally, by comparing the results listed in Table 4, Table 5 and Table 6, it can be concluded that TPIS outperforms the compared methods.

**Discussion**

In this study, TPIS is proposed for differential diagnosis of TB and pneumonia. It has two main steps for early diagnosis and final decision making, respectively. Potentially infected patients are the people acquiring low confidence scores in the first step. The final diagnosis for them will be made in the second step based on the supplementary laboratory tests, chest X-ray reports, and meta features extracted in the first step.

Table 7 illustrates the confusion matrix of the first step of TPIS by categorizing patients based on their confidence scores.

**Table 7- Confusion matrix of the first step module categorizing patients based on their confidence score**

|  | Real class: P | | | Real class: TB | | |
|---|---|---|---|---|---|---|
|  | CS=0 | CS = 0.5 | CS = 1 | CS=0 | CS = 0.5 | CS = 1 |

| | | 6% | 83% | | 1% | 1% |
|---|---|---|---|---|---|---|
| **Predicted class: P** | | | | | | |
| **Predicted class: TB** | | 5% | 2% | | 18% | 78% |
| **Suspicious patients** | 4% | | | 2% | | |

As shown by Table 7, totally, 7% of patients suffering from pneumonia and 2% of patients suffering from TB are misclassified in the first step of TPIS. Moreover, 4% of patients in class P and 2% of patients in class TB cannot be classified with majority voting because their assigned votes for two classes are the same. Their confidence score is 0.

By setting a minimum threshold for confidence scores, the potentially infected patients are ones with confidence score less than the predefined threshold. They will be instructed to pursue supplementary laboratory tests.

According to Table 7, it is recommended that the minimum threshold should be 0.51 or more. This means that if the majority or all of classifiers assigned the same label to a patient, then we will not go forward to the next step; however, if there is a strong disagreement between labels that classifiers assigned, then we will go to the second step of our model which will provide the ultimate decision. By setting the threshold equal to 0.51, 15% (6+5+4) of the potentially infected patients suffering from pneumonia and 21% (1+18+2) of the potentially infected patients suffering from TB are directed to the second diagnosis step.

Based on this policy, only 2% of the patients suffering from pneumonia and 1% of the patients with TB are misclassified in the early stage disease diagnosis; no supplementary laboratory tests are suggested. These 3% of the patients are misdiagnosed by TPIS.

For potentially infected patients, the confusion matrix of the second step of TPIS is shown in Table 8.

**Table 8- confusion matrix of final decision making for potentially infected patients in the second step of TPIS**

| | Real class: P (totally 15% of patients suffering from Pneumonia) | Real class: TB (totally 21% of patients suffering from pneumonia) |
|---|---|---|
| Real class: P | 94% | 17% |
| Real class: TB | 6% | 83% |

As shown in Table 8, in the second step, 6% from 15% of the patients suffering from Pneumonia (0.06*0.015*119 = 1) and 17% from 21% of the patients suffering from TB (0.17 * 0.21 * 80 = 3) are misclassified. It means that totally, 4 patients are misdiagnosed from 35 suspicious patients.

Moreover, in the first step, 2% of the patients suffering from pneumonia (0.02*119=2) and 1% of the patients suffering from TB (0.01*80=1) are misclassified.

By aggregating the results of the first and the second steps of TPIS, 7 patients out of 199 patients are misdiagnosed (about 3.6%). On the other hand, as listed in Table 5, if all patients are referred for laboratory testing and send their information to the second step of our DSS, the accuracy is about 93.89. However, by applying our policy to suggest laboratory testing only for suspicious patients, we will achieve the accuracy of 96.48. Our research shows that supplementary laboratory tests and X-ray reports do not improve the accuracy of disease diagnosis for all patients. In fact, our method of only testing potentially infected patients has proven to be more effective than testing all patients.

Figure 7 shows the workflow of TPIS to reduce the error rate of differential diagnosis of TB and pneumonia.

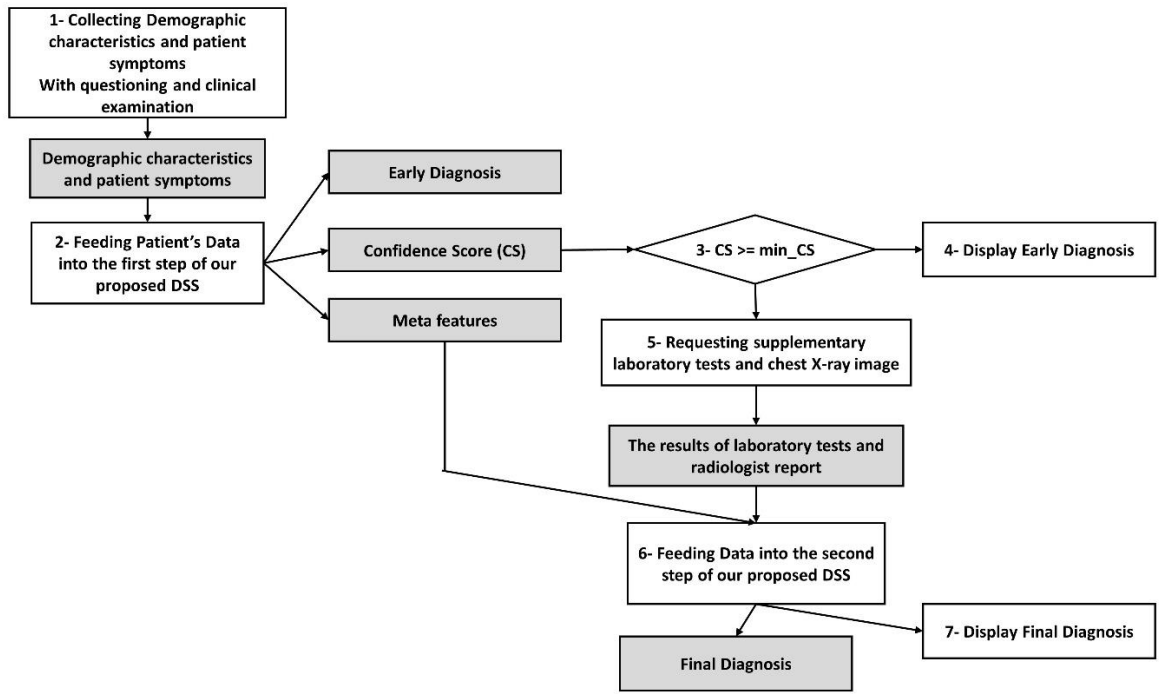

**Figure 7- the workflow of TPIS**

**Conclusions**

Tuberculosis and pneumonia have similar symptoms. Therefore, the differential diagnosis of these two diseases is a challenging task. In this study, a decision support system is proposed for distinguishing TB from pneumonia in two main steps. The first step uses a stacked ensemble classifier, which is taught the clinical symptoms and demographic characteristics of the patients. The classifier makes an early diagnosis with a confidence score. If the confidence score is lower than a predefined threshold, supplementary laboratory tests and chest X-rays are requested for final decision making in the second step of TPIS. The stacked ensemble classifier in the second step is taught the results of the laboratory tests, X-ray reports, and meta features extracted in the first step. The experimental results show that TPIS outperforms the compared methods.

The main limitations of this study and their related future work opportunities follow:

- The quality of the data gathered from the physical examination highly depends on each physician's expertise. One of the features described by physicians is the lung sound pattern. Automating lung sound analysis could be the next step in eliminating individual physicians' limitation. Extracting and adding some features describing lung sound pattern with machine learning methods can augment the collected data.
- In this study, only X-ray interpretations are used, and the images are not processed directly for extracting some valuable and differentiating descriptors. Conversely, the radiologic report accuracy is positively correlated with the radiologist's expertise. We suggest that feature extraction from X-ray images should be performed by automatic machine learning and deep learning methods. It may improve the diagnosis accuracy without depending on the radiologist's expertise.
- Another limitation of our study was the absence of smear tests in the patients' records. For future research, we propose adding a smear test and smear culture result to the data to increase the accuracy of classifying the patients' records into TB and pneumonia.
- Some features are excluded from this study because of their high rate of missing value. Using automated methods for collecting and archiving data can reduce the missing value rate of the features in the future.

Another proposed future direction is trying to differentiate pulmonary TB from other diseases having similar symptoms. Finally, collecting healthy patients' data and adding them to the study is suggested in order to differentiate healthy patients from TB or pneumonia.

**List of abbreviations:**

- AUC: Area Under ROC curve
- CI: Confidence Interval
- CRISP: Computer Retrieval of Information on Scientific Projects
- CRP: C-reactive protein
- CS: Confidence Score
- CXR: Chest X-ray
- DT: Decision Tree
- ESR: Erythrocyte Sedimentation Rate
- FP: False Positive
- FN: False Negative
- GBT: Gradient Boosting Tree
- K-NN: K- nearest neighbors
- LR: Logistics Regression
- MCV: Mean Corpuscular Volume
- N: Negative
- P: Positive
- RF: Random forest
- ROC: Receiver Operating Curve
- SE: Stacked Ensemble
- SVM: Support Vector Machines
- TB: Mycobacterium Tuberculosis
- TP: True Negative
- TPIS: Tuberculosis-Pneumonia Diagnostic Information System
- TN: True Negative
- WBC: White Blood Cell

**Declarations:**

**Ethics approval:**

The study was approved by Psychology and Counseling Organization Of Markazi Province (Arak). The ethics approval certificate is uploaded in the journal submission site as a supplementary document.

**Consent to participate:**

The consent to participate in data collection has been taken verbal by the physicians. This research is a retrospective study and the authors has gathered data for this research from paper-based patient medical records. All patients' records have been filled by the physicians and the authors are responsible for extracting data from paper-based records to excel files. It is approved in the ethics approval certificate written by Psychology and Counseling Organization Of Markazi Province (Arak).

**Consent for publication:**

" Not applicable"

**Availability of data and materials**

The data that support the findings of this study are not publicly available since the restrictions apply to the availability of these data, which were used under license for the current study.

**Competing interests:**

The authors declare that they have no competing interests.

**Funding:**

This study was not funded by any organization.

**Authors' Contribution**

Conceptualization: TK, AF and HS
Data curation: TK, AF and HS
Formal analysis: TK and AF
Funding acquisition: there is no funding.
Investigation: TK and AF
Methodology: TK and AF
Project administration: T Khatibi
Software: TK and AF
Supervision:  TK
Validation: TK, AF and HS
Visualization: TK and AF
Writing – original draft: TK and AF
Writing – review & editing: TK, AF and HS
all authors have read and approved the manuscript.

**Acknowledgements:**

Not applicable.

**Figure legends**

Figure 1- (left): the main steps of this study methodology, (right): the proposed DSS (TPIS) framework

Figure 2- histograms of some binary features

Figure 3- the framework of the first step of TPIS with more details

Figure 8- the framework of the second step of TPIS with more details

Figure 5- the evaluation framework used in this study

Figure 6- Accuracy, AUC and F-Score of the compared classifiers

Figure 7- the workflow of TPIS